\documentclass[sigconf]{acmart}

\AtBeginDocument{%
  \providecommand\BibTeX{{%
    \normalfont B\kern-0.5em{\scshape i\kern-0.25em b}\kern-0.8em\TeX}}}

\usepackage[textwidth=2cm,textsize=small]{todonotes}

\usepackage{esvect}
\usepackage[linesnumbered,ruled]{algorithm2e}
\usepackage{multirow}
\usepackage{subfigure}
\usepackage{tikz}
\usepackage{amsmath}
\usepackage{xparse}
\usepackage{balance}
\usepackage{dblfloatfix}
\usepackage{float}
\NewDocumentCommand\DownArrow{O{2.0ex} O{red}}{%
   \mathrel{\tikz[baseline] \draw [<-, line width=0.5pt, #2] (0,0) -- ++(0,#1);}
}

\NewDocumentCommand\UpArrow{O{2.0ex} O{blue}}{%
   \mathrel{\tikz[baseline] \draw [->, line width=0.5pt, #2] (0,0) -- ++(0,#1);}
}

\copyrightyear{2020}
\acmYear{2020}
\setcopyright{iw3c2w3}
\acmConference[WWW '20]{Proceedings of The Web Conference 2020}{April 20--24, 2020}{Taipei, Taiwan}
\acmBooktitle{Proceedings of The Web Conference 2020 (WWW '20), April 20--24, 2020, Taipei, Taiwan}
\acmPrice{}
\acmDOI{10.1145/3366423.3380227}
\acmISBN{978-1-4503-7023-3/20/04}

\begin{document}

\title[The POLAR Framework]{The POLAR Framework: Polar Opposites Enable Interpretability of Pre-Trained Word Embeddings}

\author{Binny Mathew}
\authornote{Both authors contributed equally to this research.}
\authornote{The work was done during internship at RWTH Aachen University}
\affiliation{%
  \institution{IIT Kharagpur, India}
  }
\email{binnymathew@iitkgp.ac.in}

\author{Sandipan Sikdar}
\authornotemark[1]
\affiliation{%
  \institution{RWTH Aachen University, Germany}
  }
\email{sandipan.sikdar@cssh.rwth-aachen.de}

\author{Florian Lemmerich}
\affiliation{%
  \institution{RWTH Aachen University, Germany}
  }
\email{florian.lemmerich@cssh.rwth-aachen.de}
 
\author{Markus Strohmaier}
\affiliation{%
  \institution{RWTH Aachen University \& GESIS, Germany}
  }
\email{markus.strohmaier@cssh.rwth-aachen.de}

\renewcommand{\shortauthors}{Mathew and Sikdar, et al.}

\begin{abstract}
 
We introduce `\emph{POLAR}' --- a framework that adds interpretability to pre-trained word embeddings via the adoption of semantic differentials. 
  Semantic differentials are a psychometric construct for measuring the semantics of a word by analysing its position on a scale between two polar opposites (e.g., cold -- hot, soft -- hard). 
 The core idea of our approach is to transform existing, pre-trained word embeddings via semantic differentials to a new ``polar'' space with interpretable dimensions defined by such polar opposites. 
 Our framework also allows for selecting the most discriminative dimensions from a set of polar dimensions provided by an oracle, i.e., an external source.  
 We demonstrate the effectiveness of our framework 
  by deploying it to various downstream tasks, in which our interpretable word embeddings achieve a performance that is comparable to the original word embeddings. 
  We also show that the interpretable dimensions selected by our framework align with human judgement.
  Together, these results demonstrate that interpretability can be added to word embeddings without compromising performance.
Our work is relevant for researchers and engineers interested in interpreting pre-trained word embeddings.   
\end{abstract}



\maketitle

\section{Introduction}
Dense distributed word representations such as Word2Vec~\cite{Mikolov:2013:DRW:2999792.2999959} and GloVe~\cite{pennington2014glove} have been established as a key step for technical solutions for a wide variety of natural language processing tasks including translation~\cite{zou2013bilingual}, sentiment analysis~\cite{socher2013recursive}, and image captioning~\cite{you2016image}. While such word representations have substantially contributed towards improving performance of such tasks, it is usually difficult for humans to make sense of them. 
At the same time, interpretability of machine learning approaches is essential for many scenarios, for example to increase trust in predictions~\cite{ribeiro2016should}, to detect potential errors, or to conform with legal regulations such as General Data Protection Regulation (GDPR~\cite{EU_GDPR_2016}) in Europe that recently established a ``right to explanation".
Since word embeddings are often crucial for downstream machine learning tasks, the non-interpretable nature of word embeddings often impairs a deeper understanding of their performance in downstream tasks.

\begin{figure}[t]
	\centering
	\includegraphics[scale=0.257]{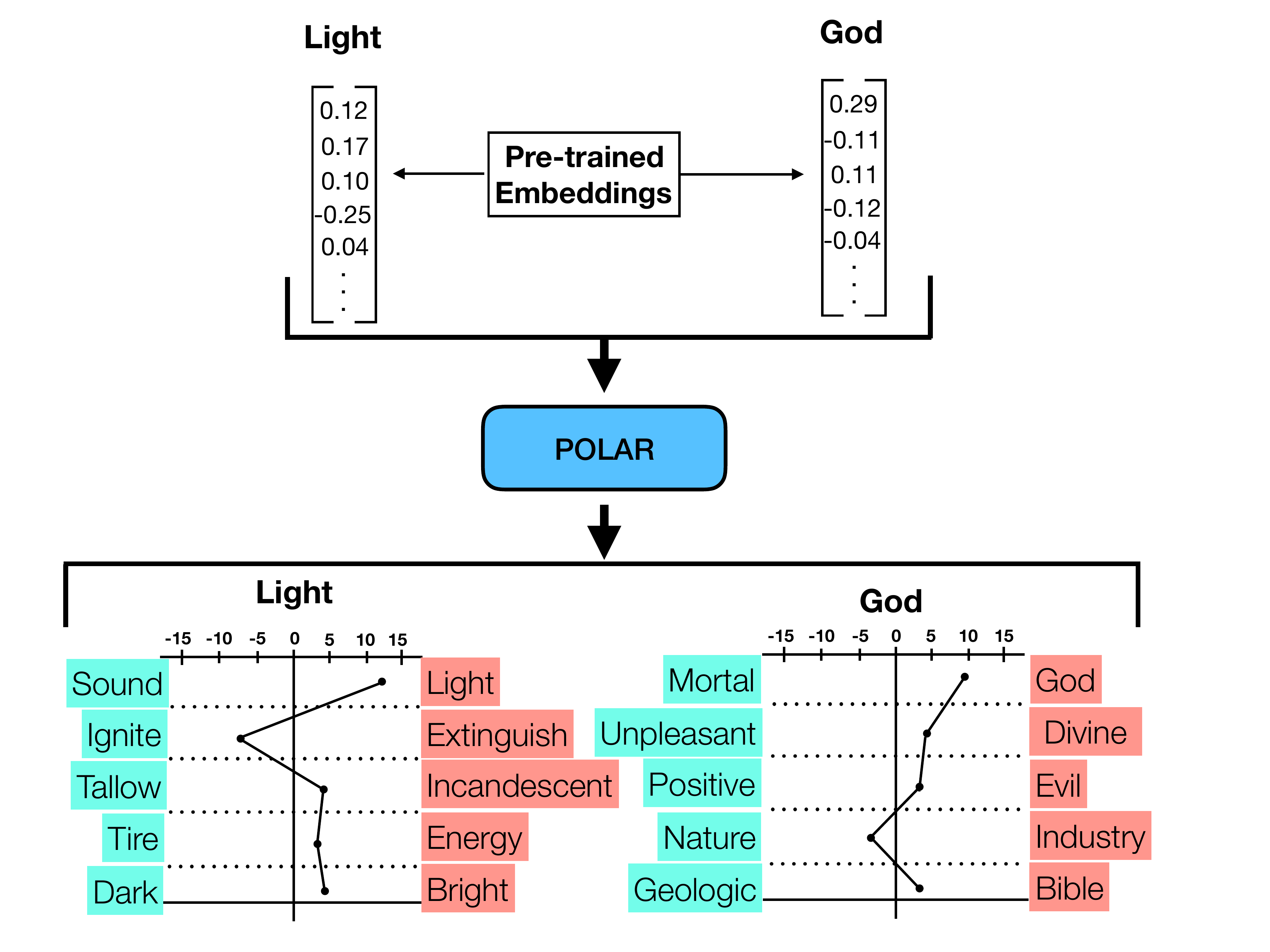}
	\caption{\emph{The POLAR Framework.} The framework takes pre-trained word embeddings as an input and generates word embeddings with interpretable (polar) dimensions as an output. In this example, the embeddings are generated by applying POLAR to embeddings pre-trained on Google News dataset with Word2Vec.}
	\label{fig:illust}
\end{figure}

\noindent{\bf Problem}.
We aim to add interpretability to an arbitrarily given pre-trained word embedding via post-processing in order make embedding dimensions interpretable for humans (an illustrative example is provided in figure~\ref{fig:illust}). 
Our objective is explicitly \emph{not improving} performance per se, but adding interpretability while \emph{maintaining} performance on downstream tasks.

\noindent{\bf Approach}. The POLAR framework utilizes the idea of semantic differentials (\citet{osgood1957measurement}) that allows for capturing connotative meanings associated with words and applies it to word embeddings. 
To obtain embeddings with interpretable dimensions, we first take a set of polar opposites from an oracle (e.g., from a lexical database such as WordNet), and identify the corresponding polar subspace from the original embedding. The basis vectors of this polar subspace are calculated using the vector differences of the polar opposites.
The pre-trained word vectors are then projected to this new polar subspace, which enables the interpretation of the transformed vectors in terms of the chosen polar opposite pairs.
Because the set of polar opposites could be potentially very large, we also discuss and compare several variations to select expressive subsets of polar opposite pairs to use as basis for the new vector space.

\noindent{\bf Results and contribution}. We evaluate our approach with regard to both \emph{performance} and \emph{interpretability}.  
With respect to performance, we compare the original embeddings with the proposed POLAR embeddings in a variety of downstream tasks. We find that in all cases the performance of POLAR embeddings is competitive with the original embeddings. In fact, for a few tasks POLAR even outperforms the original embeddings.
Additionally, we evaluate interpretability with a human judgement experiment. We observe that in most cases, but not always, the dimensions deemed as most discriminative by POLAR, align with dimensions that appear most relevant to humans. 
Our results are robust across different embedding algorithms. 
This demonstrates that we can augment word embeddings with interpretability without much loss of performance across a range of tasks.

To the best of our knowledge, our work is the first to apply the idea of semantic differentials - stemming from the domain of psychometrics - to word embeddings.
Our POLAR framework provides two main advantages: (i) It is agnostic w.r.t. the underlying model used for obtaining the word vectors, i.e., it works with arbitrary word embedding frameworks such as GloVe and Word2Vec. (ii) as a post-processing step for pre-trained embeddings, it neither requires expensive (re-)training nor access to the original textual corpus. 
Thus, POLAR enables the addition of interpretability to arbitrary word embeddings \emph{post-hoc}. 
To facilitate reproducibility of our work and enable their use in practical applications, we make our implementation of the approach publicly available\footnote{Code: \url{https://github.com/Sandipan99/POLAR}}.

\section{Background}
In this section, we provide a brief overview of prior work on interpretable word embeddings as well as the semantic differential technique pioneered by Osgood.

\subsection{Interpretable word embeddings}

One of the major issues with low-dimensional dense vectors utilized by Word2Vec~\cite{Mikolov:2013:DRW:2999792.2999959} or GloVe~\cite{pennington2014glove} is that the generated embeddings are difficult to interpret. Although the utility of these methods has been demonstrated in many downstream tasks, the meaning associated with each dimension is typically unknown. To solve this, there have been few attempts to introduce some sense of interpretability to these embeddings~\cite{murphy2012learning,panigrahi2019Word2Sense,subramanian2018spine,faruqui2015sparse}.

Several recent efforts have attempted to introduce interpretability by making embeddings sparse. In that regard, Murphy et al. proposed to use a Non-Negative Sparse Embedding (NNSE) in order to to obtain sparse and interpretable word embeddings~\cite{murphy2012learning}.
\citet{fyshe2014interpretable} introduce a joint Non-Negative Sparse Embedding (JNNSE) model to capture brain activation records along with texts. The joint model is able to capture word semantics better than text based models. \citet{faruqui2015sparse} transform the dense word vectors derived from Word2Vec using sparse coding (SC) and demonstrate that the resulting word vectors are more similar to the interpretable features used in NLP. However, SC usually suffers from heavy memory usage since it requires a global matrix. This makes it quite difficult to train SC on large-scale text data. To tackle this, \citet{luo2015online} propose an online learning of interpretable word embeddings from streaming text data. \citet{sun2016sparse} also use an online optimization algorithm for regularized stochastic learning which makes the learning process efficient. This allows the method to scale up to very large corpus.

\begin{figure}[t]
	\centering
	\includegraphics[width=0.8\linewidth]{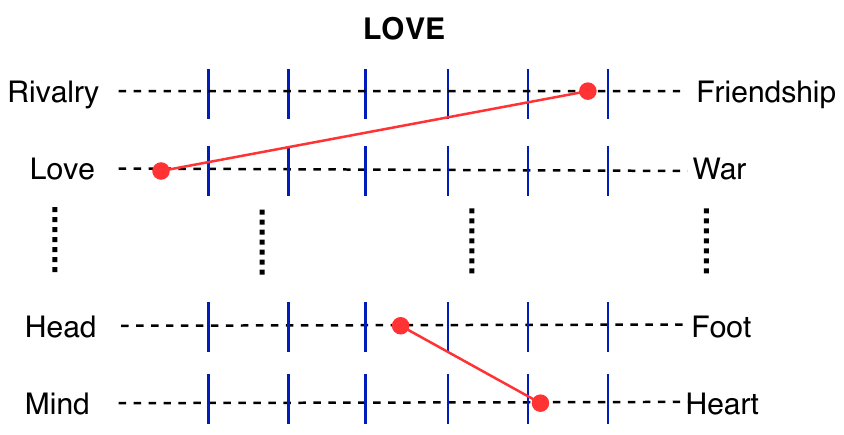}
	\caption{\emph{An example semantic differential scale.} The example reports the response of an individual to the word \textit{Love}. Each dimension represents a semantically polar pair. A response close to the edge means a strong relation with the dimension and a response near the middle means no clear relation. }
	\label{fig:semantic_scale}
\end{figure}

\citet{subramanian2018spine} utilize  denoising $k$-sparse autoencoder to generate efficient and interpretable distributed word representations. 
The work by \citet{panigrahi2019Word2Sense} is to the best our knowledge, among the existing research, closest to our work. 
The authors propose \textit{Word2Sense} word embeddings in which each dimension of the embedding space corresponds to a fine-grained sense, and the non-negative value of the embedding along a dimension represents the relevance of the sense to the word. Word2Sense is a generative model which recovers senses of a word from the corpus itself. However, these methods would not be applicable if the user does not have access to the corpus itself. Also, such models have high computation costs, which might make it infeasible for many users who wish to add interpretability to word embeddings.

Our work differs from the existing literature in several ways. The existing literature does not necessarily provide dimensions that are actually interpretable to humans in an intuitive way. By contrast, our method represents each dimension as a pair of polar opposites given by an oracle (typically end users, a dictionary, or some vocabulary), which assigns direct meaning to a dimension. 
Moreover, massive computation costs associated with training these models have led researchers to adopt pre-trained embeddings for their tasks. The proposed POLAR framework, being built on top of pre-trained embeddings and not requiring the corpus itself, suits this common design.

\subsection{Semantic Differentials}
The semantic differential technique by \citet{osgood1957measurement} is used to measure the connotative meaning of abstract concepts. 
This scale is based on the presumption that a concept can have different dimensions associated with it, such as the property of speed, or the property of being good or bad. The semantic differential technique is meant for obtaining a 
person's psychological reactions to certain concepts, such as persons or ideas, under study. It consists of a number of bipolar words that are associated with a scale. The survey participant indicates an attitude or opinion by checking on any one of seven spaces between the two extremes of each scale. For an example, consider Figure ~\ref{fig:semantic_scale}. Here, each dimension of the scale represents a semantically polar pair such as `Rivalry' and `Friendship', `Mind' and `Heart'. A participant could be given a word (such as `Love') and asked to select points along each dimension, which would represent his/her perception of the word. A point closer to the edge would represent a higher degree of agreement with the concept. 
The abstract nature of semantic differential allows it to be used in a wide variety of scenarios. Often, antonym pairs are used as polar opposites. For example, this is related to work by \citet{an2018semaxis},  in which the authors utilize polar opposites as semantic axes to generate domain-specific lexicons as well as capturing semantic differences in two corpora. 
It is also similar to the tag genome (\citet{vig2012tag}), a concept that is used to elicit user preferences in tag-based systems. 

Overall, the semantic differential scale is a well established and widely used technique for observing and measuring the meaning of concepts such as information system satisfaction (\citet{xue2011punishment}), attitude toward information technology (\citet{bhattacherjee2004understanding}) information systems planning success (\citet{doherty1999relative}), perceived enjoyment (\citet{luo2011web}), or website performance (\citet{huang2005web}).

This paper brings together two isolated concepts: word embeddings and semantic differentials. We propose and demonstrate that the latter can be used to add interpretability to the former.

\section{Methodology}
\label{sec:Method}
In this section, we introduce the POLAR Framework and elaborate how it generates interpretable word embeddings. Note that we do not train the used word embeddings from scratch rather we generate them by post-processing embeddings already trained on a corpus.

\begin{figure*}[htpb]
	\begin{center}
		\subfigure[POLAR overview]{\label{flow_diag}\includegraphics[scale=0.34]{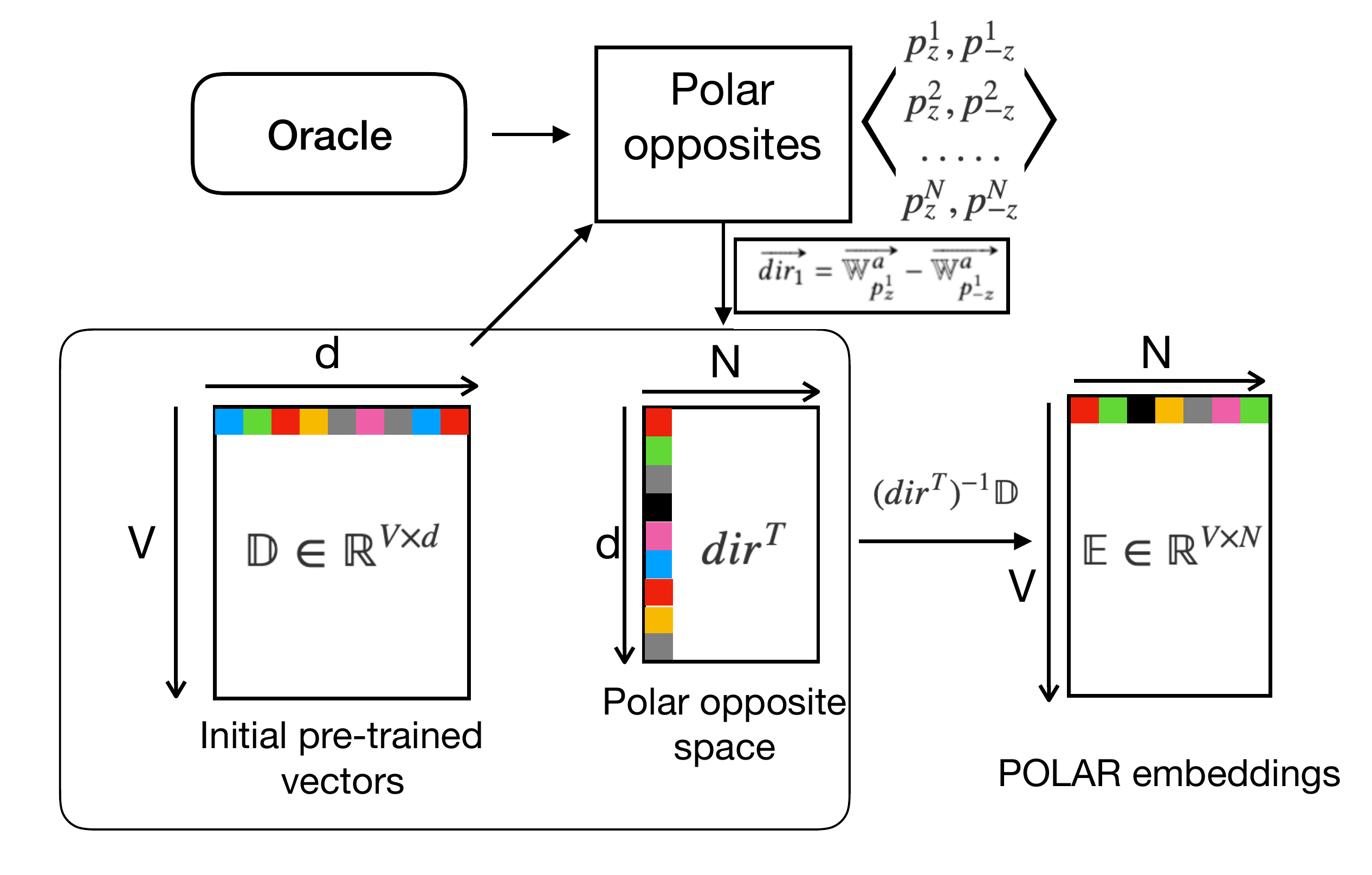}}
		\subfigure[POLAR transformation]{\label{fig:method_1}\includegraphics[scale=0.245]{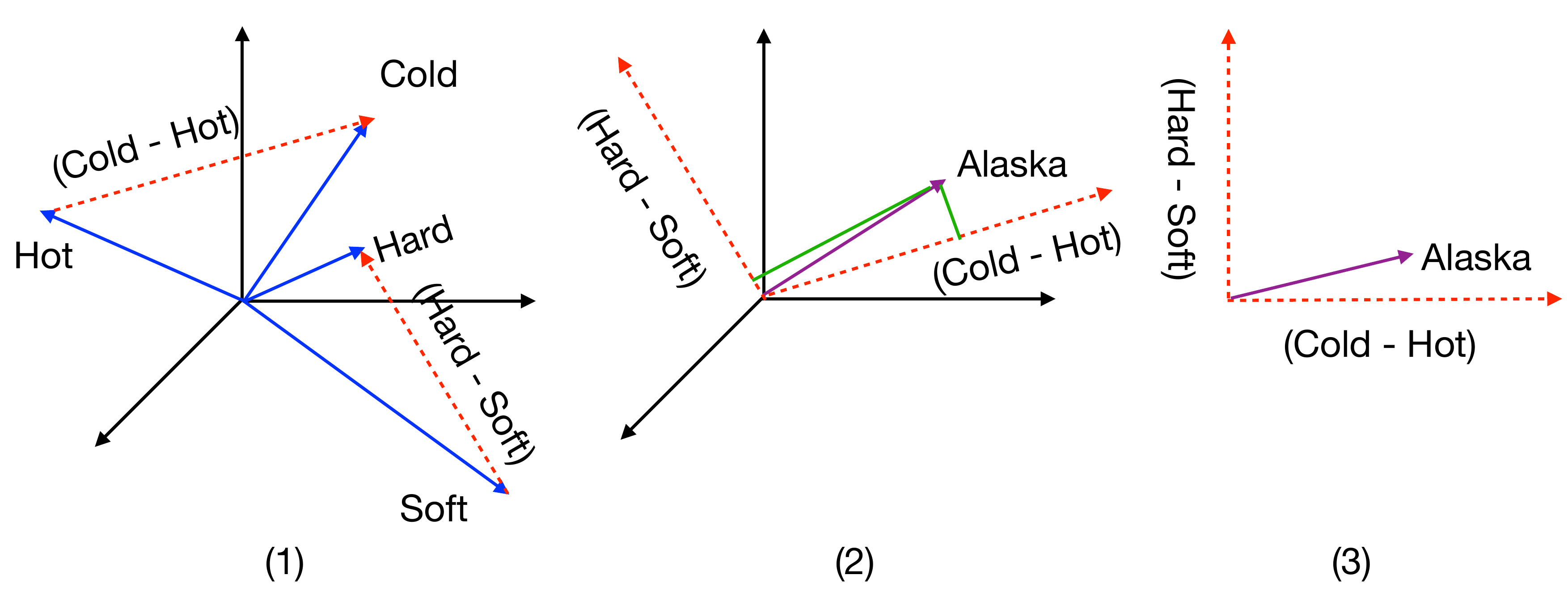}}
	\end{center}
	\caption{\label{fig:method} (a) \emph{Visual illustration of the POLAR framework}. A set of pre-trained embeddings ($\mathbf{R}^{V\times d}$) represents the input to our approach, and we assume that an Oracle provides us with a list of polar opposites with which we generate the polar opposite space ($\mathbf{R}^{d\times N}$). We apply change of basis transform to obtain the final embeddings ($\mathbf{R}^{V\times N}$). Note that $V$ is the size of the vocabulary, $N$ is the number of polar opposites and $d$ is the dimension of the pre-trained embeddings. (b) \emph{POLAR transformation.} In this example the original size of the embeddings is three and we consider two polar opposites (cold, hot) and (hard, soft). In the first step (1) we obtain the direction of the polar opposites (vectors in the original space represented in \textcolor{blue}{blue}) which also represent the change of basis vectors for the polar subspace (represented by \textcolor{red}{red} dashed lines). In the second step (2) we project the original word vectors (`Alaska' in this case) to this polar subspace. After the transformation, `Alaska' gets aligned more to the (cold, hot) direction which is much more related to `Alaska' than the (hard, soft) direction (3).}
\end{figure*}

\subsection{The POLAR framework}
Consider a corpus with vocabulary $\mathcal{V}$ containing $V$ words.
For each word $v \in \mathcal{V}$, the corresponding embedding trained using an algorithm $a$ (Word2Vec, GLoVE) is denoted by $\vv{\mathbb{W}_{v}^{a}} \in \mathbb{R}^d$, where $d$ denotes the dimension of the embedding vectors. In this setting, let $\mathbb{D} = [\vv{\mathbb{W}_1^a}, \vv{\mathbb{W}_2^a}, \vv{\mathbb{W}_3^a}, \dots, \vv{\mathbb{W}_V^a}] \in \mathbb{R}^{V \times d}$ denote the set of pretrained word embeddings which is used as input to the \emph{POLAR} framework. Note that $\vv{\mathbb{W}_i^a}$ is a unit vector with $||\vv{\mathbb{W}_i^a}|| = 1$.

The key idea is to identify an interpretable subspace and then project the embeddings to this new subspace in order to obtain interpretable dimensions which we call \emph{POLAR dimensions}. To obtain this subspace we consider a set of $N$ polar opposites. In this paper, we use a set of antonyms (e.g., hot--cold, soft--hard etc.) as an initial set of polar opposites, but this could easily be changed to arbitrary other polar dimensions. Typically, we assume that these set of polar opposites are provided by some \emph{oracle}, i.e., an external source that provides polar, interpretable word pairs.

Given these set of $N$ polar opposites, we now proceed to generate the polar opposite subspace. Let the set of polar opposites be denoted by $\mathbb{P} = \{ (p_z^{1}, p_{-z}^{1}), (p_z^{2}, p_{-z}^{2}), \dots ,(p_z^{N}, p_{-z}^{N})\}$. Now the direction of a particular polar opposite $(p_z^{1}, p_{-z}^{1})$ can be obtained by:
\begin{align}
\vv{dir_1} = \vv{\mathbb{W}_{p_z^1}^a} - \vv{\mathbb{W}_{p_{-z}^1}^a}
\end{align}

The direction vectors are calculated across all the polar opposites and stacked to obtain $dir \in \mathbb{R}^{N \times d}$.  
Note that $dir$ represents the change of basis matrix for this new (polar) embedding subspace $\mathbb{E}$. Let a word $v$ in the embedding subspace $\vv{\mathbb{E}}$ be denoted by $\vv{\mathbb{E}_v}$. So for $v$ we have by the rules of linear transformation:
\begin{align}
dir^T \, \vv{\mathbb{E}_v} &= \vv{\mathbb{W}_v^a}\\
\vv{\mathbb{E}_v} &= (dir^T)^{-1} \, \vv{\mathbb{W}_v^a}\, 
\end{align}

Note that each dimension (\emph{POLAR dimension}) in this new space $\vv{\mathbb{E}}$, can be interpreted in terms of the polar opposites. 
The inverse of the matrix $dir$ is accomplished through Moore-Penrose generalized inverse~\cite{ben2003generalized} usually represented by $dir^{+}$. 
While this can in most settings be computed quickly and reliably, there is one issue: when the number of polar opposites (i.e., \emph{POLAR dimensions}), is similar to the number of dimensions of the original embedding, the change of basis matrix $dir$ becomes ill conditioned and hence the transformed vector $\vv{\mathbb{E}_v}$ becomes meaningless and unreliable. We discuss this in more detail later in this paper. 
Note that with $N$ polar opposites, the worst case complexity of calculating the generalized inverse is $O(N^3)$. Since $N  \ll V$ and the inverse needing to be calculated just once, the computation is overall very fast (e.g. $< 5$ seconds for $N = 1,468$). Performance can further be improved using parallel architecture ($0.29$ seconds on a $48$ core machine).
The overall architecture of the model is presented in Figure~\ref{flow_diag}.

We illustrate using a toy example in Figure \ref{fig:method_1}. In this setting the embeddings trained on a corpus are of dimension $d$. The polar opposites $\mathbb{P}$ in this case are (hard',`soft') and (`cold',`hot'). In the first step, we obtain the direction of the polar opposites, which is then followed by projecting the words (`Alaska' in this example) into this new subspace. After the transformation, `Alaska' gets aligned more to the (cold--hot) direction which is much more related to `Alaska' than the (hard--soft) direction.
\begin{table*}[bp]
	\caption{\emph{Performance of POLAR across different downstream tasks.} We compare the embeddings generated by POLAR against the initial Word2Vec and GloVe vectors on a suite of benchmark downstream tasks. For all the tasks, we report the accuracy when using the original embeddings vis-a-vis when using POLAR embeddings (the classification model is same in both cases). In all the tasks we achieve comparable results with POLAR for both Word2Vec and GloVe (we report the percentage change in performance as well). In fact, for Religious News classification and Question classification we perform better than the original embeddings trained on Word2Vec.}
	\begin{tabular}{l|l||cc||cc}
		\hline
		\multicolumn{2}{l||}{Tasks}                               & Word2Vec & Word2Vec w/ POLAR & GloVe & GloVe w/ POLAR \\ \hline
		\multirow{3}{*}{News Classification}      & Sports        &   $0.947$       &   \(\displaystyle 0.922 \DownArrow \textcolor{red}{2.6\%} \)              &  $0.951$     &  0.951 $\updownarrow$ $0.0\%$           \\ \cline{2-6} 
		& Religion      & $0.812$         &    \(\displaystyle 0.849 \UpArrow \textcolor{blue}{4.6\%} \)              &  $0.876$     &   \(\displaystyle 0.852 \DownArrow \textcolor{red}{2.7\%} \)           \\ \cline{2-6} 
		& Computers     & $0.737$         &   \(\displaystyle 0.717 \DownArrow \textcolor{red}{2.7\%} \)              &  $0.804$     &  \(\displaystyle 0.802  \DownArrow \textcolor{red}{0.2\%} \)            \\ \hline
		\multicolumn{2}{l||}{Noun Phrase Bracketing}              &     $0.792$     &     \(\displaystyle 0.761 \DownArrow \textcolor{red}{3.9\%} \)            &    $0.764$   &   \(\displaystyle 0.757    \DownArrow \textcolor{red}{0.9\%} \)           \\ \hline
		\multicolumn{2}{l||}{Question Classification}             & $0.954$         &    \(\displaystyle 0.958 \UpArrow \textcolor{blue}{0.4\%} \)             & $0.962$      & \(\displaystyle 0.964   \UpArrow \textcolor{blue}{0.2\%} \)             \\ \hline
		\multicolumn{2}{l||}{Capturing Discriminative Attributes} &     $0.639$     &    \(\displaystyle 0.628 \DownArrow \textcolor{red}{1.7\%} \)             &  $0.633$     &   \(\displaystyle 0.638    \UpArrow \textcolor{blue}{0.7\%} \)           \\ \hline
		\multicolumn{2}{l||}{Word Analogy}                        &     $0.740$     &    \(\displaystyle 0.704 \DownArrow \textcolor{red}{4.8\%} \)             &    $0.751$   &   \(\displaystyle 0.727  \DownArrow \textcolor{red}{3.1\%} \)           \\ \hline
		\multicolumn{2}{l||}{Sentiment Classification}            &   $0.816$       &  \(\displaystyle 0.821 \UpArrow \textcolor{blue}{0.6\%} \)               & $0.808$      &   \(\displaystyle 0.818 \UpArrow \textcolor{blue}{1.2\%} \)            \\ \hline
	\end{tabular}
	\label{tab:downstream_tasks}
\end{table*}

While in our explanations we only use antonyms, polar opposites could also include other terms, such as political terms representing politically opposite ideologies (e.g. republican vs. democrat) that could be obtained from political experts, or people representing opposite views (e.g. Chomsky vs. Norvig) that could be obtained from domain experts.  

\subsection{Selecting POLAR dimensions}
\label{sec:dim_select}
We also design a set of algorithms to select suitable dimensions as POLAR embeddings from a larger set of candidate pairs of polar opposites. For all the algorithms, we use the same notation with $P$ denoting the initial set of polar opposite vectors ($|P| = N$) and $O$ denoting the reduced set of polar opposite  vectors (initialized to $\phi$) obtained utilizing the algorithms discussed below. $K$ denotes the specified size of $O$. 

\noindent{\bf Random selection.} In this simple method, we randomly sample $K$ polar opposite vectors from $P$ and add them to $O$. For experimental evaluation, We repeat this procedure with different randomly selected sets and report the mean value across runs.

\noindent{\bf Variance maximization.} 
In this method, we select the dimensions (polar opposite vectors) based on the value of their variance on the vocabulary. 
Typically, for each dimension, we consider the value corresponding to each word in the vocabulary when projected on it and then calculate the variance of these values across each dimension.  
We take the top $K$ polar opposite vectors (POLAR dimensions) from $P$ which have the highest value of variance and add them to $O$. This is motivated by the idea that the polar opposites with maximum variance encode maximum information.\\

\noindent{\bf Orthogonality maximization.}
The primary idea here is to select a subset of polar opposites in such a way that the corresponding vectors are maximally orthogonal.
Typically, we follow a greedy approach to generate the subset of polar vectors as presented in Algorithm \ref{algo:dimension_selection}. 
First, we obtain a vector with maximum variance (as in Variance maximization) from $P$ and add it to $O$. In each of the following steps we subsequently add a vector to $O$ such that it is maximally orthogonal to the ones that are already in $O$.
A candidate vector $z$ at any step is selected via -

\begin{align}
z = \underset{x \in P}{\mathrm{argmin}} \frac{1}{|O|} \sum_{n=1}^{n=|O|} \vv{O_i}\cdot\vv{x} 
\end{align}

We then continue the process until a specified number of dimensions $K$ is reached.

\IncMargin{0.5em}
\begin{algorithm}
	\SetKwInOut{Input}{Input}\SetKwInOut{Output}{Output}
	
	\Input{
		$P$ -- Initial set of polar opposite vectors, $K$ -- the required size}
	\Output{$O$ -- The reduced set of polar opposite vectors consisting of $K$ vectors}
	$O \gets \emptyset$\;
	$U \gets$ Select a vector from $P$ with maximum variance\;
	$O \gets O \cup {U} $\;
	$P \gets P-U $  //Remove $U$ from $P$\;
	
	\For{$i \gets 2$ \textbf{to} $K$}{
		$min\_vec \gets \emptyset$\;
		$min\_score \gets +\infty$\;
		\ForEach{$curr\_vec \in P$} {
			$curr\_score \gets$ Average\_Score($P$,$ curr\_vec$)\;
			\If{$curr\_score < min\_score$} {
				$min\_score \gets curr_score$\;
				$min\_vec \gets curr\_vec$\;
			}
			$O \gets O \cup {min\_vec} $\;
			$P \gets P-min\_vec $  //Remove $U$ from $P$\;
		}
	}
	\Return{$O$}\;
	
	\caption{Orthogonality maximization}
	\label{algo:dimension_selection}
\end{algorithm}

\section{Experimental Setup}

Next, we discuss our experimental setup including details on the used polar opposites, training models, and baseline embeddings. As our framework does not require any raw textual corpus for embedding generation, we use two popular pretrained embeddings:
\begin{enumerate}
	\item {\bf Word2Vec} embeddings~\cite{Mikolov:2013:DRW:2999792.2999959}\footnote{\url{https://drive.google.com/file/d/0B7XkCwpI5KDYNlNUTTlSS21pQmM}} trained on Google News dataset. The model consists of $3$ million words with an embedding dimension of $300$.
	\item {\bf GloVe} embeddings~\cite{pennington2014glove}\footnote{We used the Common Crawl embeddings with 42B tokens: \url{https://nlp.stanford.edu/projects/glove/}} trained on Web data from Common Crawl. The model consists of $1.9$ million words with embedding dimension set at $300$. 
\end{enumerate}

As \textbf{polar opposites} we adopt the antonym pairs used in previous literature by \citet{shwartz2017hypernyms}\footnote{The datasets are available here: \url{https://github.com/vered1986/UnsupervisedHypernymy/tree/master/datasets}}. These antonym pairs were collected from the Lenci/Benotto Dataset~\cite{santus2014unsupervised} as well as the EVALution Dataset~\cite{santus2015evalution}. The antonyms in both datasets were combined to obtain a total of 4,192 antonym pairs. After this, we removed duplicates to get 1,468 unique antonym pairs. In the following experiments, we will be using these 1,468 antonym pairs to generate POLAR embeddings\footnote{ In case, a word in the antonym pair is absent from the Word2Vec/GloVe vocabulary, we ignore that pair. Ergo, we have 1,468 pairs for GloVe but 1,465 for Word2Vec.}. However, we study the effect of size of the embeddings on different downstream tasks later in this paper. 
It is important to reiterate at this point that we do not intend to improve the performance of the original word embeddings. Rather we intend to add interpretability \emph{without much loss in performance} in downstream tasks.  

\section{Evaluation of Performance}

We follow the same procedure as in \citet{faruqui2015sparse}, \citet{subramanian2018spine}\footnote{We use the evaluation code given in \url{https://github.com/harsh19/SPINE}}, and \citet{panigrahi2019Word2Sense} to evaluate the performance of our method on downstream tasks. We use the embeddings in the following downstream classification tasks: news classification, noun phrase bracketing, question classification, capturing discriminative attributes, word analogy, sentiment classification and word similarity. In all these experiments we use the original word embeddings as baseline and compare their performance with POLAR-interpretable word embeddings.

\begin{table*}[bp]
	\centering
	\caption{\emph{Performance of POLAR on word similarity evaluation across multiple datasets.} Similarity between a word pair is measured by human annotated scores as well as cosine similarity between the word vectors. 
		For each dataset, we report the spearman rank correlation $\rho$ between the word pairs ranked by human annotated score as well as the cosine similarity scores (we report the percentage change in performance as well). POLAR consistently outperforms the baseline original  embeddings (refer to Table~\ref{tab:word_similarity_task}) in case of GloVe while in case of Word2Vec, the performance of POLAR is comparable to the baseline original embeddings for most of the datasets. }
	\begin{tabular}{c|l||cc||cc}
		\hline
		Task & Dataset    & Word2Vec & Word2Vec w/ POLAR & GloVe & Glove w/ POLAR  \\ \hline\hline
		\multirow{9}{*}{Word Similarity}
		&Simlex-999 &0.442          &\(\displaystyle 0.433  ~\DownArrow \textcolor{red}{2.0\%} \)    &0.374     &\(\displaystyle 0.455 ~\UpArrow  \textcolor{blue}{21.7\%} \)                \\ \cline{2-6}
		&WS353-S    &0.772          &\(\displaystyle 0.758  ~\DownArrow \textcolor{red}{1.8\%} \)   &0.695      &\(\displaystyle 0.777 ~\UpArrow  \textcolor{blue}{11.8\%} \)                \\ \cline{2-6}
		&WS353-R    &0.635          &\(\displaystyle 0.554  \DownArrow \textcolor{red}{12.8\%} \)  &0.600      &\(\displaystyle 0.683 ~\UpArrow  \textcolor{blue}{13.8\%} \)                \\ \cline{2-6}
		&WS353      &0.700          &\(\displaystyle 0.643  ~\DownArrow \textcolor{red}{8.1\%} \)  &0.646       &\(\displaystyle 0.733 ~\UpArrow  \textcolor{blue}{13.5\%} \)                \\ \cline{2-6}
		&MC         &0.800          &\(\displaystyle 0.789  ~\DownArrow \textcolor{red}{1.4\%} \)  &0.786       &\(\displaystyle 0.869  ~\UpArrow  \textcolor{blue}{10.6\%} \)               \\ \cline{2-6}
		&RG         &0.760          &\(\displaystyle 0.764  ~\UpArrow  \textcolor{blue}{0.5\%} \)  &0.817       &\(\displaystyle 0.808 ~\DownArrow  \textcolor{red}{1.1\%} \)                \\ \cline{2-6}
		&MEN        &0.771          &\(\displaystyle 0.761  ~\DownArrow \textcolor{red}{1.3\%} \) &0.736        &\(\displaystyle 0.783 ~\UpArrow  \textcolor{blue}{6.4\%} \)                \\ \cline{2-6}
		&RW         &0.534          &\(\displaystyle 0.484  ~\DownArrow \textcolor{red}{9.4\%} \) &0.384        &\(\displaystyle 0.451 \UpArrow  \textcolor{blue}{17.5\%} \)                \\ \cline{2-6}
		&MT-771     &0.671          &\(\displaystyle 0.659  ~\DownArrow \textcolor{red}{1.8\%} \) &0.684        &\(\displaystyle 0.678 ~\DownArrow  \textcolor{red}{0.9\%} \)                \\ \hline
	\end{tabular}
	
	\label{tab:word_similarity_task}
\end{table*}

\subsection{News Classification}
As proposed in \citet{panigrahi2019Word2Sense}, we consider three binary classification tasks from the 20 news-groups dataset\footnote{http://qwone.com/~jason/20Newsgroups/}.\\
\noindent{\bf Task.} 
Overall the dataset consists of three classes of news articles: (a) sports, (b) religion and (c) computer. For the `sports' class, the task involves a binary classification problem of categorizing an article to `baseball' or `hockey' with training/validation/test splits ($958$/$239$/$796$). For `religion', the classification problem involves `atheism' vs. `christian' ($870$/$209$/$717$) while for `computer' it involves `IBM' vs. `Mac' ($929$/$239$/$777$).\\
\noindent{\bf Method.} Given a news article, a corresponding feature vector is obtained by averaging over the vectors of the words in the document. We use a wide range of classifiers including support vector classifiers (SVC), logistic regression, random forest classifiers for training and report the test accuracy for the model which provides the best validation accuracy. \\
\noindent{\bf Result.} We report  a comparison of classification accuracies between classifiers with the original embeddings vs. those with POLAR interpretable embeddings in Table \ref{tab:downstream_tasks} for the three tasks. For Word2Vec embeddings, POLAR performs almost as good as the original embeddings in all the cases. In fact, the accuracy improves with POLAR for `religion' classification by $4.5\%$. We achieve similar performance with GloVe embeddings as well.\\

\subsection{Noun phrase bracketing}
\noindent{\bf Task.} The task involves classifying noun phrases as left bracketed or right bracketed. For example, given the noun phrase \emph{blood pressure medicine}, the task is to decide whether it is \{\emph{(blood pressure) medicine}\} (left) or \{\emph{blood (pressure medicine)}\} (right). 
We use the dataset proposed in~\citet{lazaridou2013fish} which constructed the Noun phrase bracketing dataset from the penn tree bank\cite{marcus1993building} that consists of $2,227$ noun phrases with three words each.

\noindent{\bf Method.} Given a noun phrase, we obtain the feature vector by averaging over the vectors of the words in the phrase. We use SVC (with both linear and RBF kernel), Random forest classifier and logistic regression for the task and use the model with the best validation accuracy for testing.\\
\noindent{\bf Result.} We report the accuracy score in Table~\ref{tab:downstream_tasks}. In both Word2Vec and GloVe, we obtain similar results when using POLAR instead of the corresponding original vectors ($0.792$, $0.761$). The results are even closer in case of GloVe ($0.764$, $0.757$).

\subsection{Question Classification}
\noindent{\bf Task.} The question classification task \cite{li2002learning} involves classifying a question into six different types, e.g., whether the question is about a location, about a person or about some numeric information. The training dataset contains $5,452$ labeled questions, and the test dataset consists of 500 questions. By  isolating  10\%  of  the  training  questions for validation, we use train/validation/test splits of 4,906/546/500 questions respectively.\\
\noindent{\bf Method}. As in previous tasks, we create feature vectors for a question by averaging over the word vectors of the constituent words. We train with different classification models (SVC, random forest and logistic regression) and report the best accuracy across the trained models.\\ 
\noindent{\bf Result.}
From Table \ref{tab:downstream_tasks} we can see that POLAR embeddings are able to marginally outperform both Word2Vec ($0.954$ vs. $0.958$) and GloVe embeddings ($0.962$ vs. $0.964$).

\subsection{Capturing Discriminative Attributes}
\noindent{\bf Task}. The Capturing Discriminative Attributes task (\citet{krebs2018semeval}) was introduced at SemEval 2018. The aim of this task is to identify whether an attribute could help discriminate between two concepts. For example, a successful system should determine that \textit{red} is a discriminating attribute in the concept pair \textit{apple, banana}. The purpose of the task is to better evaluate the capabilities of state-of-the-art semantic models, beyond pure semantic similarity. It is a binary classification task on the dataset\footnote{The dataset is available here: \url{https://github.com/dpaperno/DiscriminAtt}} with training/validation/test splits of 17,501/2,722/2,340 instances. The dataset consists of triplets of the form (\textit{concept1, concept2, attribute}).

\noindent{\bf Method}. We used the unsupervised distributed vector cosine baseline as suggested in \citet{krebs2018semeval}. The main idea is that the discriminative attribute should be close to the word it characterizes and farther from the other concept. If the cosine similarity of \textit{concept1} and \textit{attribute} is greater than the cosine similarity of \textit{concept2} and \textit{attribute}, we say that the \textit{attribute} is discriminative.

\noindent{\bf Result.} We report the accuracy in Table~\ref{tab:downstream_tasks}. We achieve comparable performance when using POLAR embeddings instead of the original ones. In fact accuracy is slightly better in case of GloVe.

\begin{figure*}[!tb]
	\begin{center}
		\subfigure[Sports News classification]{\label{fig:ncSpor}\includegraphics[scale=0.15]{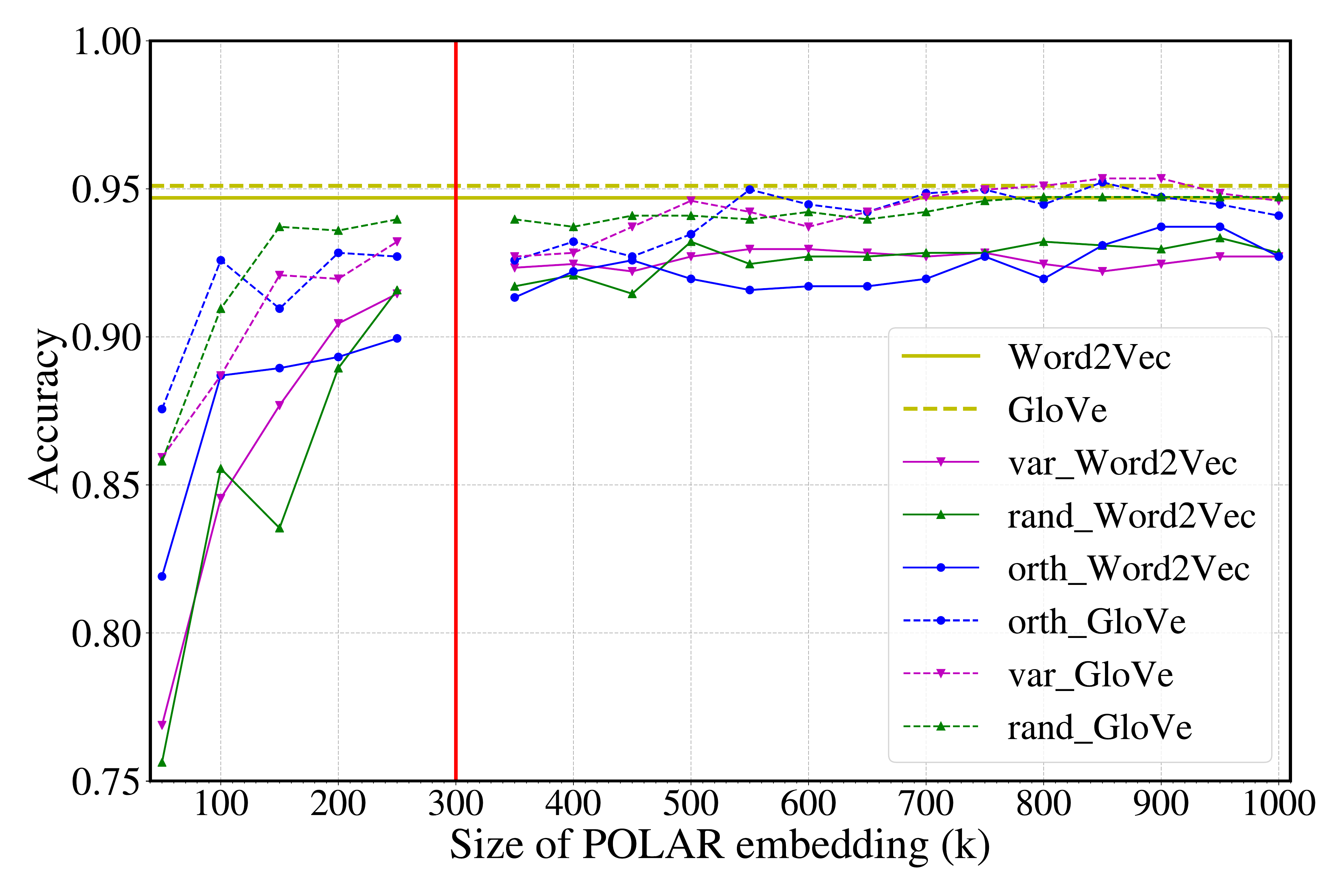}}
		\subfigure[Religion News classification]{\label{fig:ncRel}\includegraphics[scale=0.15]{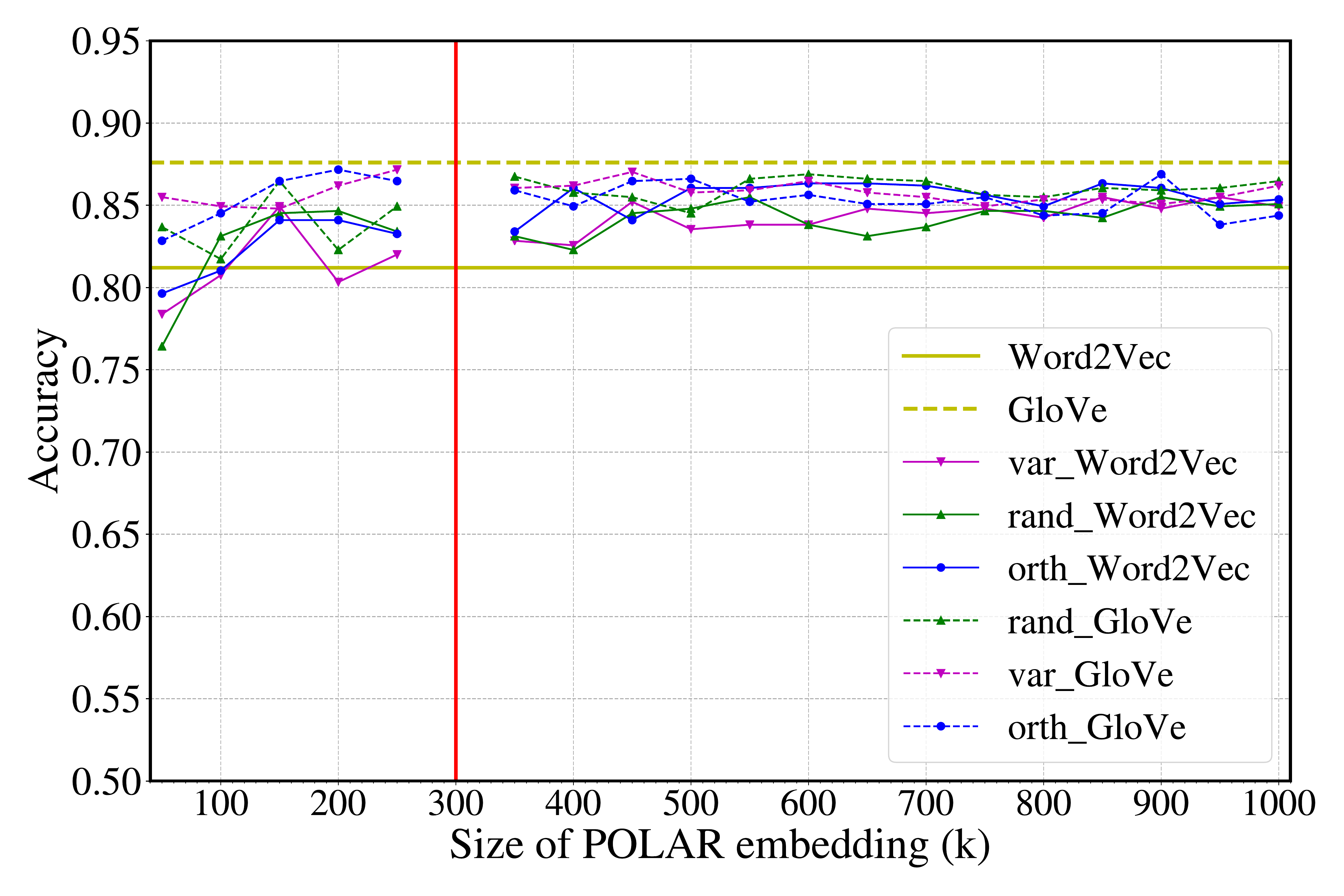}}
		\subfigure[Computers News classification]{\label{fig:ncComp}\includegraphics[scale=0.15]{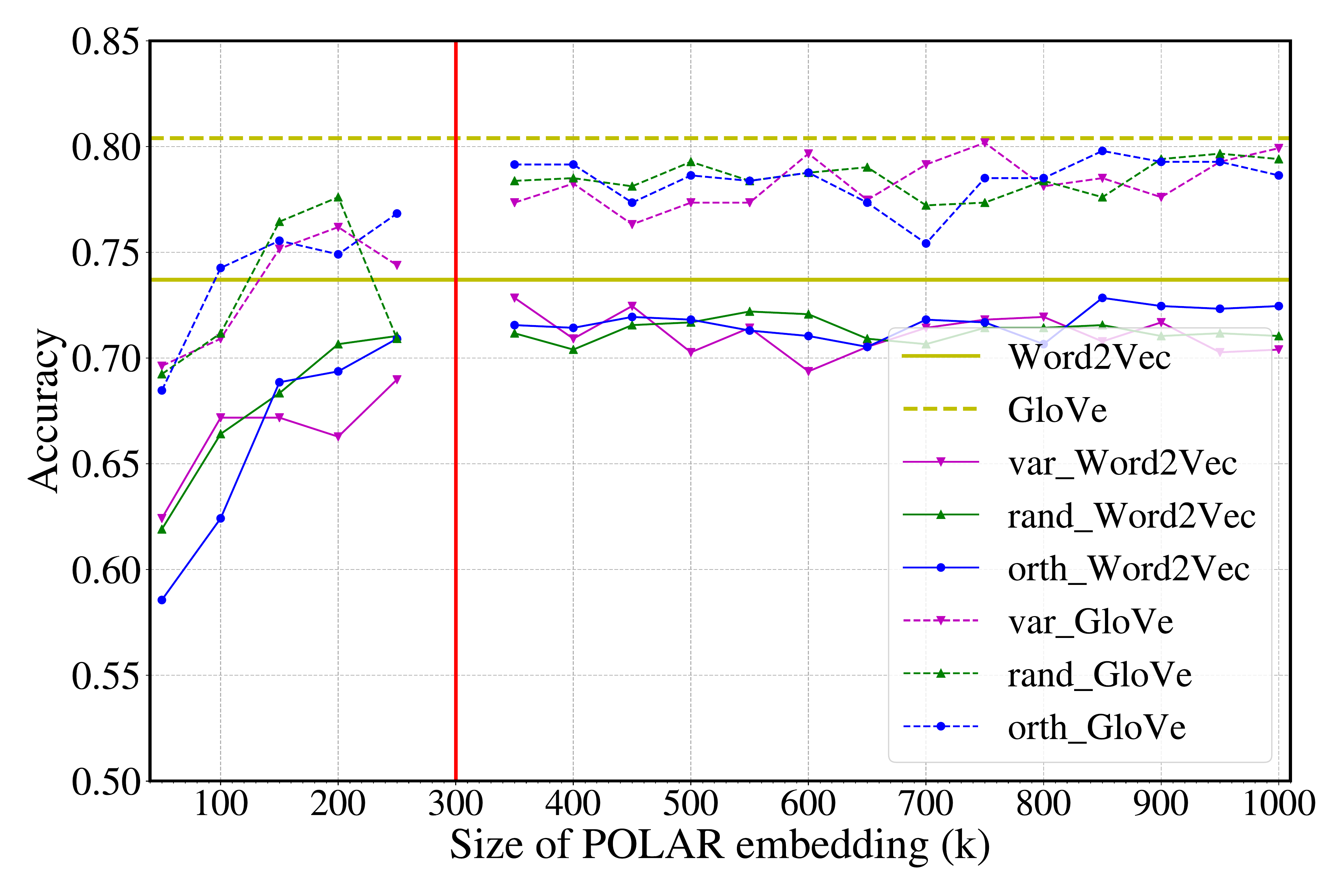}}\\
		\subfigure[Noun phrase bracketing]{\label{fig:npbrack}\includegraphics[scale=0.15]{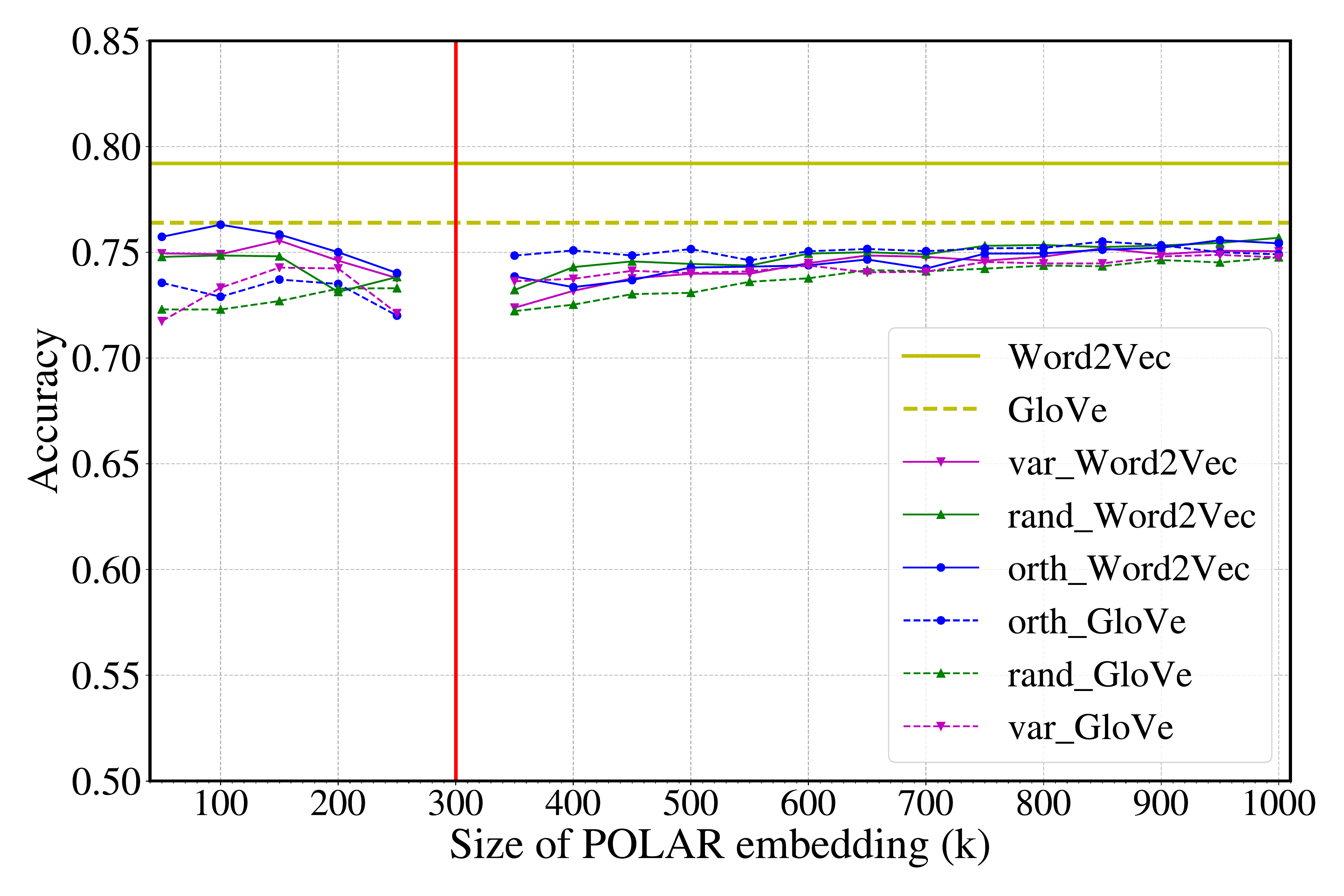}}
		\subfigure[Question classification]{\label{fig:TREC}\includegraphics[scale=0.15]{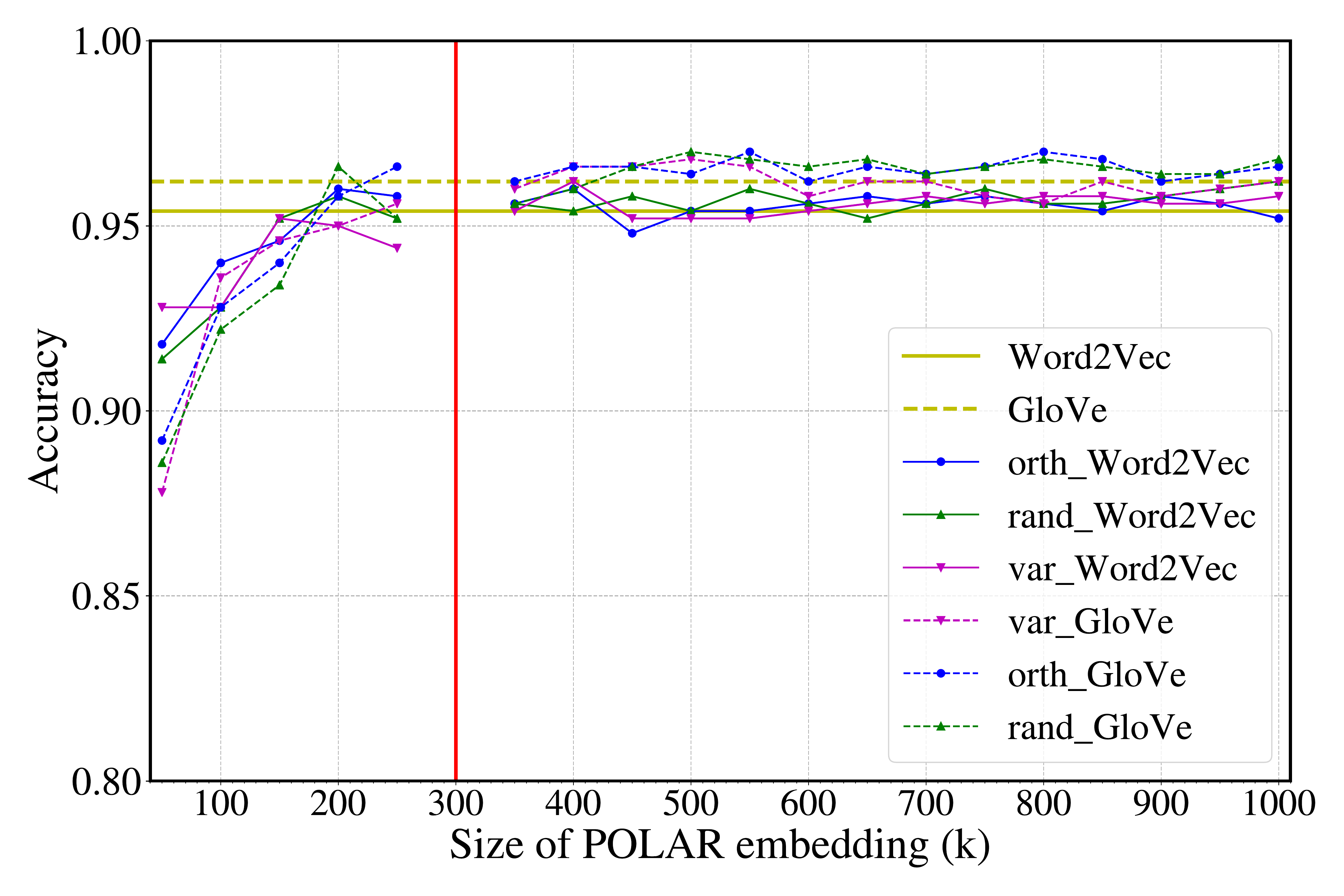}}
		\subfigure[Capturing discriminative attributes]{\label{fig:disc}\includegraphics[scale=0.15]{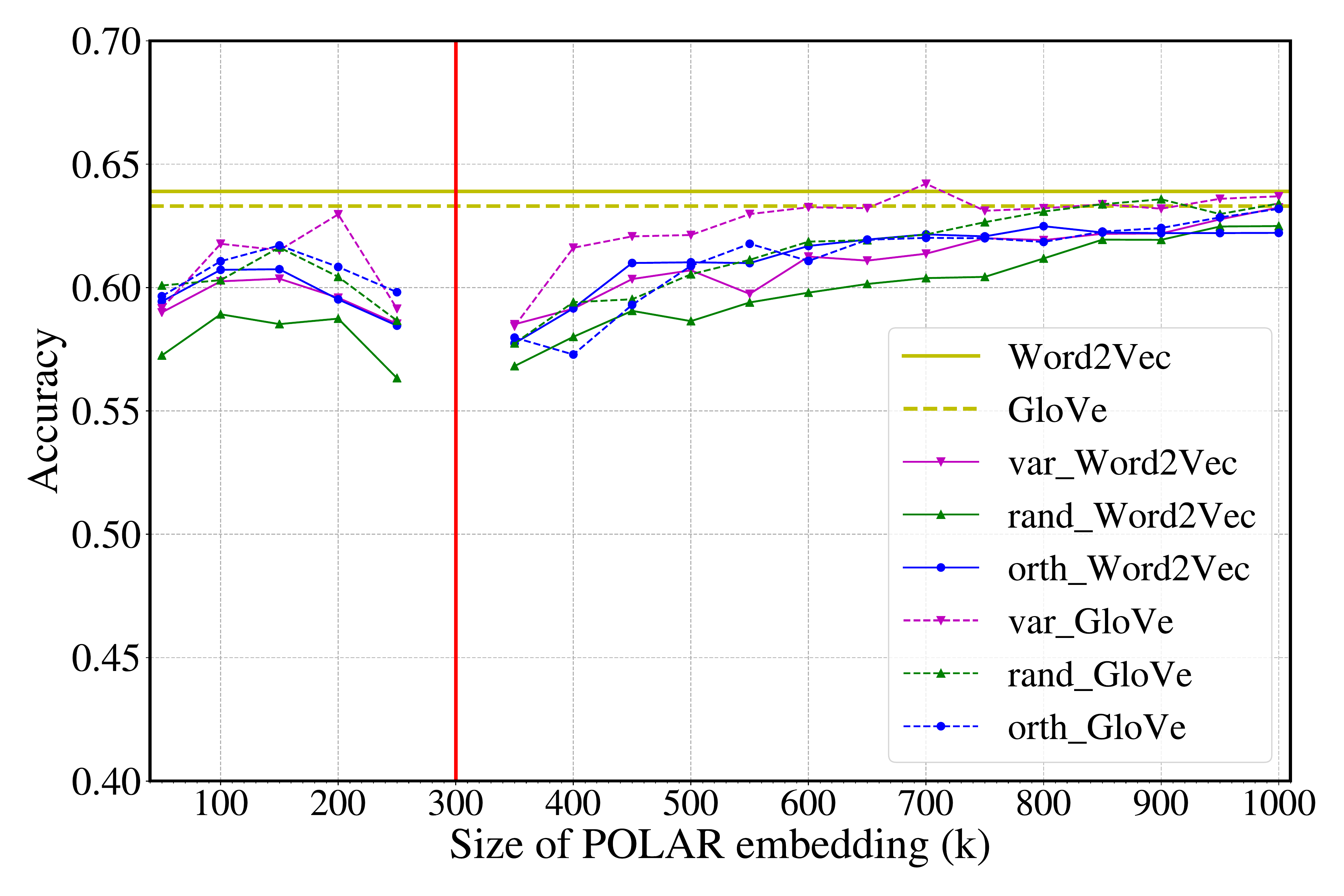}}
		
	\end{center}
	\caption{\emph{Dependency on embedding size}. We report the accuracy of POLAR as well as the original embeddings for different downstream tasks for varying sizes ($k$) of the embeddings. The dimensions are selected using three strategies - 1. random (rand), 2. maximizing orthogonality (orth) and 3. maximizing variance (var). We also report the accuracy obtained using the original Word2Vec and GloVe embeddings. 
		Although the performance improves as the embedding size increases, comparable performance is achieved with a dimension size of $200$. However, when the dimension size approaches $300$ (the dimension of the pre-trained embeddings), the change of basis vector becomes ill-conditioned and the embeddings become unreliable. We hence intentionally leave this region from the plots.}
	\label{fig:dimension}
\end{figure*}

\subsection{Word Analogy}

\noindent{\bf Task.} The word analogy task was introduced by Mikolov et al. [2013c; 2013a] to quantitatively evaluate the models' ability of encoding the linguistic regularities between word pairs. The dataset contains 5 types of semantic analogies and 9 types of syntactic analogies. The semantic analogy subset contains 8,869 questions, typically about places and people, like ``Athens is to Greece as X (Paris) is to France'', while the syntactic analogy subset contains 10,675 questions, mostly focusing on the morphemes of adjective or verb tense, such as ``run is to running as walk to walking''. \\
\noindent{\bf Method}. Word analogy tasks are typically performed using vector arithmetic (e.g.  `France + Athens - Greece') and finding the word closest to the resulting vector. We use the Gensim~\cite{rehurek_lrec}\footnote{https://radimrehurek.com/gensim/models/keyedvectors.html} to evaluate the word analogy task. \\
\noindent{\bf Result.} We achieve comparable (although not quite as good) performances with POLAR embeddings, seecTable~\ref{tab:downstream_tasks}). The performance is comparatively better in case of GloVe. 

\subsection{Sentiment Analysis}
\noindent{\bf Task.} The sentiment analysis task involves classifying a given sentence into a positive or a negative class. We utilize the Stanford Sentiment Treebank dataset~\cite{socher2013recursive} which consists of train, validation and test splits of sizes 6,920, 872 and 1,821 sentences respectively. \\
\noindent{\bf Method.} Given a sentence, the features are generated by averaging the embeddings of the constituent words. We use different classification models for training and report the best test accuracy across all the trained models.\\
\noindent{\bf Result.} We report the accuracy in Table~\ref{tab:downstream_tasks}. We achieve comparable performance when using POLAR embeddings instead of the original ones. In fact accuracy is slightly better in case of both GloVe and Word2Vec.

\subsection{Word Similarity}

\noindent{\bf Task.} The word similarity or relatedness task aims to capture the similarity between a pair of words. In this paper, we use  Simlex- 999 (~\citet{hill2015simlex}), WS353-S and WS353- R (\citet{finkelstein2002placing}), MC (~\citet{miller1991contextual}), RG (~\citet{rubenstein1965contextual}), MEN (~\citet{bruni2014multimodal}), RW (~\citet{luong2013better}) and MT-771 (~\citet{radinsky2011word,halawi2012large}). Each pair of words in these datasets is annotated by a human generated similarity score. \\
\noindent{\bf Method}. For each dataset, we first rank the word pairs using the human annotated similarity score. 
We now use the cosine similarity between the embeddings of each pair of words and rank the pairs based on this similarity score.  
Finally, we report the Spearman's rank correlation coefficient $\rho$ between the ranked list of human scores and the embedding-based rank list. Note that we consider only those pairs of words where both words are present in our vocabulary.\\
\noindent{\bf Result.} We can observe that POLAR consistently outperforms the baseline original  embeddings (refer to Table~\ref{tab:word_similarity_task}) in case of GloVe. In case of Word2Vec, the performance of POLAR is almost as good as the baseline original embeddings for most of the datasets.  

\subsection{Sensitivity to parameters}

\subsubsection{Effect of POLAR dimensions} 
In Table~\ref{tab:downstream_tasks} and Table~\ref{tab:word_similarity_task}, we report the performance of POLAR when using $1,468$ dimensions (i.e., antonym pairs). Additionally, we studied in detail the effects of POLAR dimension size on performance across the downstream tasks. As mentioned in section~\ref{sec:dim_select}, we utilize three strategies for dimension selection: (i) maximal orthogonality, (ii) maximal variance and (iii) random. 
In Figure~\ref{fig:dimension} and Figure \ref{fig:dimension_1}, we report the accuracy across all the downstream tasks for different POLAR dimensions across the three dimension selection strategies. We also report the performance of the original pre-trained embeddings in the same figures.
Typically, we observe an increasing trend i.e., the accuracy improves with increasing POLAR dimensions. But, competitive performance is achieved with $400$ dimensions for most of the tasks (even lesser for Sports and Religion News classification, Noun phrase bracketing, Question classification and Sentiment classification). 

However, a numerical inconvenience occurs when the size of POLAR dimensions approaches the dimension of the pre-trained embeddings ($300$ in our case). In this event, the columns of the change of basis matrix $dir$ loses the linear independence property making it ill-conditioned for the computation of the inverse. Hence the transformed vector of a word $v$, $\vv{\mathbb{E}_v} = (dir^T)^{-1} \, \vv{\mathbb{W}_v^a}$ (with the pseudo inverse $dir^{+} = (dir^{\ast}dir)^{-1}\,dir^{\ast}$, $dir^{\ast}$ denotes Hermitian transpose), is meaningless and unreliable. We hence eliminate the region surrounding this critical value ($300$ in this case) from our dimension related experiments (cf. Figure~\ref{fig:dimension} and Figure~\ref{fig:dimension_1}). We believe this to be a minor inconvenience as comparable results can be obtained with lower POLAR dimensions and even better with higher ones. Nevertheless, there are several regularization techniques available for finding meaningful  solutions~\cite{neumaier1998solving} for critical cases. However, exploring these techniques is beyond the scope of this paper. 

We would further like to point out that while dimension reduction is useful for comparing performance, it's not always useful to reduce the dimension itself. As argued in \citet{murphy2012learning}, interpretability often results in sparse representations and representations should model a wide range of features in the data.  
One of the main advantages of our method is it's flexibility. It can be made very sparse to capture a large array of meaning. It can also have low dimensions, and still be interpretable which is ideal for low resource corpora.

\subsubsection{Effect of the pre-trained model}

Note that we have considered embeddings trained with both Word2Vec and GloVe. The results are consistent across both the models (refer to tables ~\ref{tab:downstream_tasks} and ~\ref{tab:word_similarity_task}). This goes to show that POLAR is agnostic w.r.t the underlying training model i.e., it works across specific word embedding frameworks. Furthermore, the embeddings are trained on different corpora (Google News dataset for Word2Vec and Web data from Common Crawl in case of GloVe). This demonstrates the POLAR should work irrespective of the underlying corpora.

\subsubsection{Effect of dimension selection}

\begin{figure*}[!tb]
	\begin{center}
		\subfigure[Word analogy]{\label{fig:wa}\includegraphics[scale=0.15]{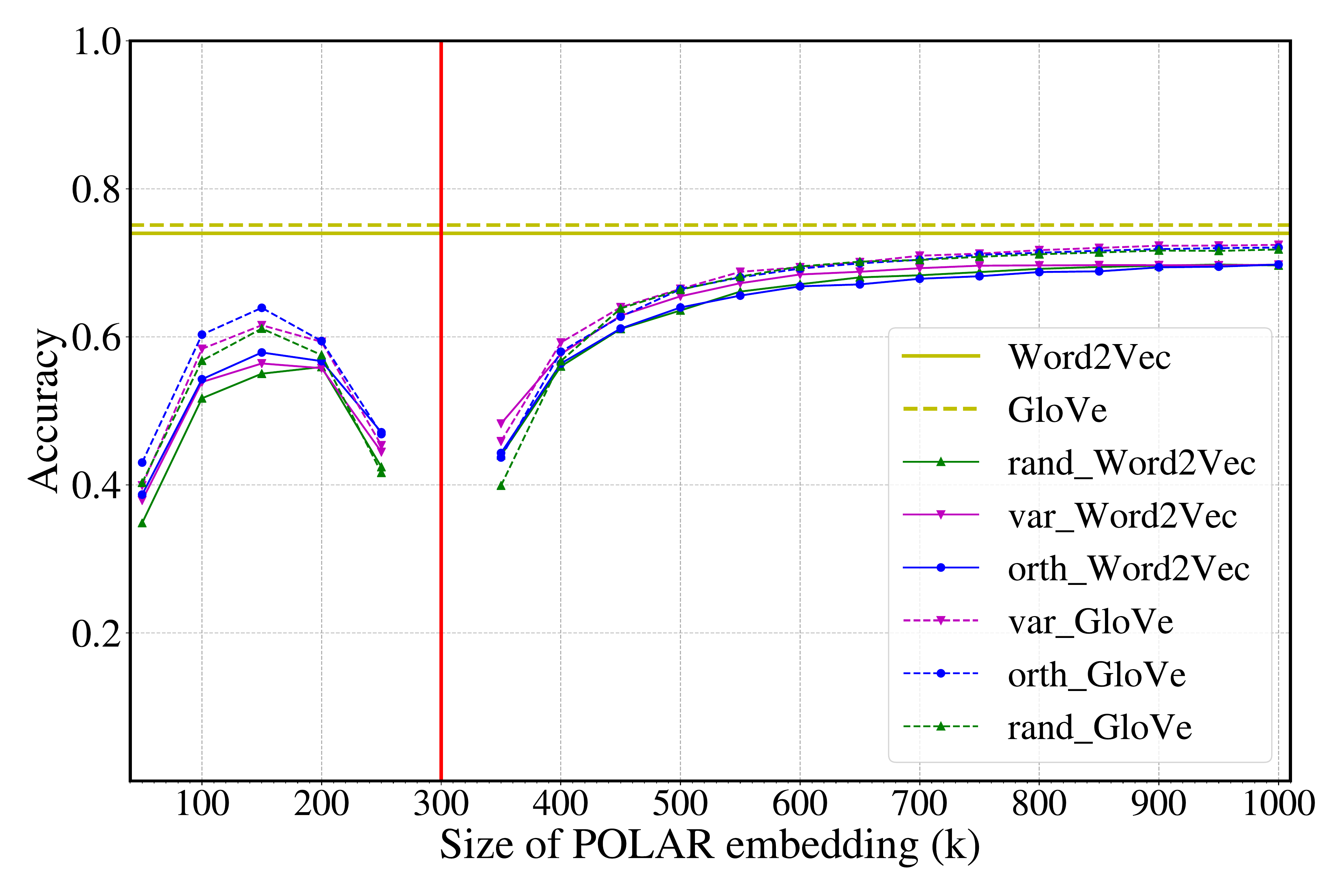}}\hspace{15mm}
		\subfigure[Sentiment classification]{\label{fig:sent}\includegraphics[scale=0.15]{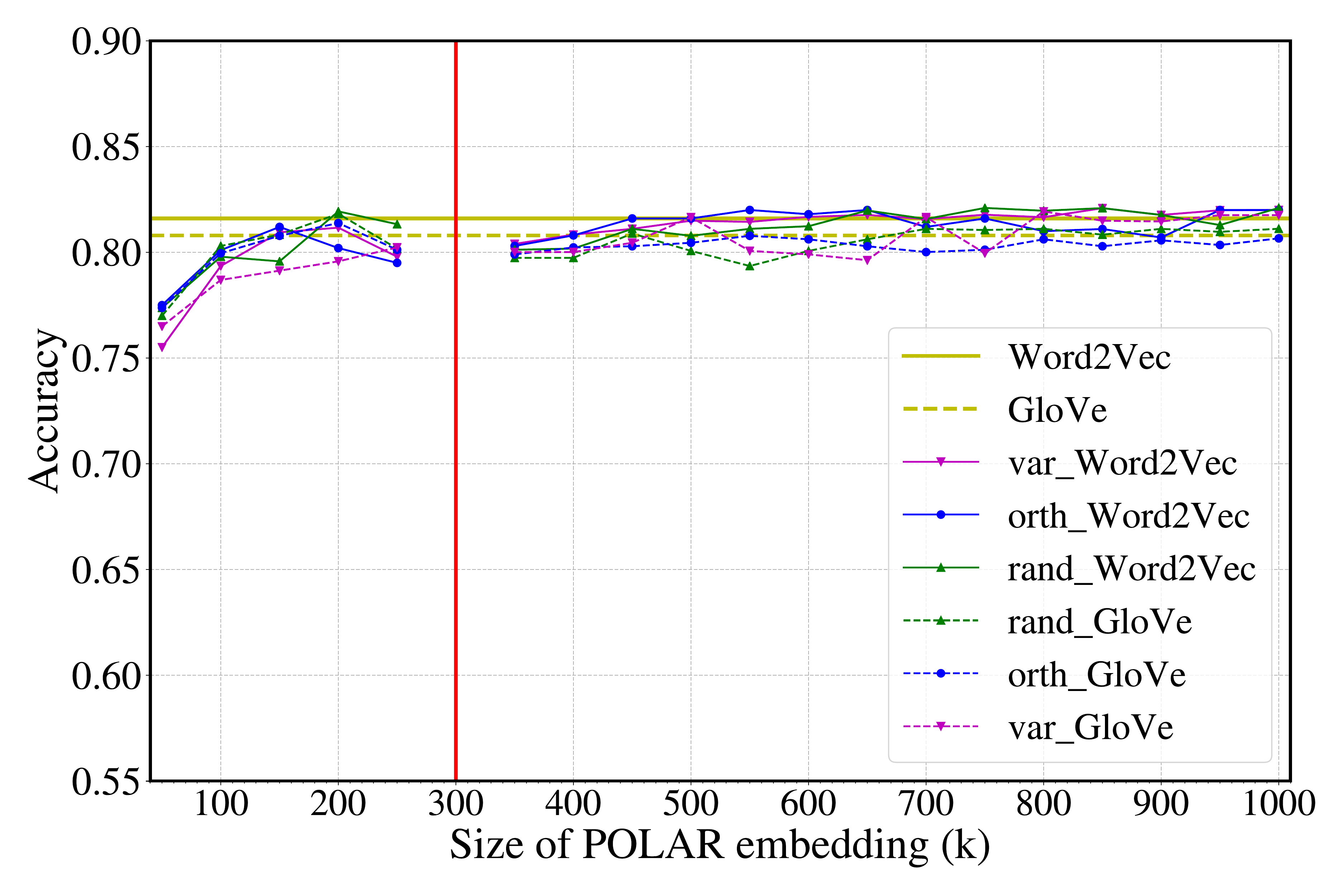}}
		
	\end{center}
	\caption{\emph{Dependency on embedding size}. We report the accuracy of POLAR in (a) word analogy tasks and (b) sentiment classification task for different sizes of the embeddings. For both tasks, the performance improves with embedding size. For word analogy comparable results are obtained at dimensions close to 600 while for sentiment classification it is around 200. 
		Owing to unreliability of the results, we leave out the results around the region of dimension size $300$.}
	\label{fig:dimension_1}
\end{figure*}

Assuming that the number of polar opposites could be large, we have proposed three methods for selecting (reducing) dimensions. 
Results presented in figures~\ref{fig:dimension} and \ref{fig:dimension_1} allow us to compare the effectiveness of these methods. Typically, we observe that all these methods have similar performances except in lower dimensions where orthogonal and variance maximization seem to perform better than random. For higher dimensions, the performances are similar.

\begin{table*}[bp]
	\centering
	\caption{\emph{Evaluation of interpretability.} The top 5 dimensions of each word using Word2Vec transformed POLAR Embedding. Note that our model is able to capture multiple interpretations of the words. Furthermore, the dimensions identified by our model are easy for humans to understand as well.}
	\begin{tabular}{r|l||r|l||r|l||r|l||r|l}
		\multicolumn{2}{c}{\hspace{-8mm}\textbf{Phone}} & \multicolumn{2}{c}{\textbf{Apple}} & \multicolumn{2}{c}{\textbf{Star}} & \multicolumn{2}{c}{\hspace{-8mm}\textbf{Cool}} & \multicolumn{2}{c}{\hspace{-8mm}\textbf{run}}  \\ \hline
		Mobile   & Stationary     & Apple       & Orange        & Actor       & Cameraman          & Cool & Geek    & Run & Stop  \\
		Fix      & Science        & Touch       & Vision       & Psychology      & Reality   & Naughty     & Nice   & Flight & Walk     \\
		Ear      & Eye            & Look        & Touch         & Sky   & Water        & Fight  & Nice              & Race & Slow \\
		Solo     & Symphonic      & Mobile      & Stationary   & Darken        & Twinkle          & Freeze   & Heat  & Organized & Unstructured         \\
		Dumb     & Philosophical  & Company     & Loneliness      & Sea      & Sky        & Add & Take  & Labor & Machine \\ \hline      
	\end{tabular}
	\vspace{-3mm}
	\label{tab:top_dimension_words}
\end{table*}

\section{Evaluation of Interpretability}
Next, we evaluate the interpretability of the dimensions produced by the POLAR framework. 

\subsection{Qualitative Evaluation}
\label{sec:in_q}
As an initial step, we sample a few arbitrary words from the embedding and transform them to POLAR embeddings using their Word2Vec representation. Based on the absolute value across the POLAR dimensions, we obtain the top five dimensions for each of these words.
In Table~\ref{tab:top_dimension_words}, we report the top 5 dimensions for these words. We can observe that the top dimensions have high semantic similarity with the word. 
Furthermore, our method is able to capture multiple interpretations of the words. This demonstrates that POLAR 
seems to be able to produce interpretable dimensions which are easy for humans to recognize.

\subsection{Human Judgement}

In order to assess the interpretability of the embeddings generated by our method, we design a human judgement experiment. For that purpose, we first select a set of 100 words randomly, considering only words with proper noun, verb, and adjective POS tags to make the comparison meaningful.

For each word, we sort the dimensions based on their absolute value and select the top five POLAR dimensions (see Section~\ref{sec:in_q} for details) to represent the word.  Additionally, we  select five dimensions randomly from the bottom 50\% from the sorted dimensions according to their polar relevance. These ten dimension are then shown as options to three human annotators each in random order with the task of selecting the five dimension which to him/her best characterize the target word. 
The experiment was performed on GloVe with POLAR embeddings. 

For each word, we assign each dimension a score depending on the number of annotators who found it relevant and select the top five (we call these ground truth dimensions). We now compare this with the top 5 dimensions obtained using POLAR. 
In Table~\ref{tab:human_judgement_topN} we report the conditional probability of the top $k$ dimensions in the ground truth to be also in POLAR. This conditional probability essentially measures, given the annotator has selected top $k$ dimensions, what is the probability that they are also the ones selected by polar or simply put, in what fraction of cases the top $k$ dimensions overlap with the Polar dimensions. 
In the same table we also note the random chance probabilities of the ground truth  dimensions to be among the POLAR dimensions (e.g., the top dimension ($k=1$) selected by the annotators, has a random chance probability of $0.5$ to be also among the POLAR dimensions). We observe that probabilities for POLAR to be much higher than random chance for all values of $k$ (refer to table~\ref{tab:human_judgement_topN}).  
In fact, the top two dimensions selected by POLAR are very much aligned to human judgement, achieving a high overlap of $0.87$ and $0.67$. On the other hand, the remaining 3 dimensions, although much better than random chance, do not reflect human judgement well. 
To delve deeper into it, we compared the responses of the 3 annotators for each word and obtained the average overlap in dimensions among them. We observed that on average the annotators agree mostly on 2-3 (mean = 2.4) dimensions (which also match with the ones selected by POLAR) but tend to differ for the rest. This goes to show that once we move out of the top 2-3 dimensions, human judgement becomes very subjective and hence difficult for any model to match.  
We interpret this as POLAR being able to capture the most important dimensions well, but unable to match more subjective dimensions in many scenarios.

\subsection{Explainable classification}
\begin{figure}[!t]
	\centering
	\includegraphics[scale=0.4]{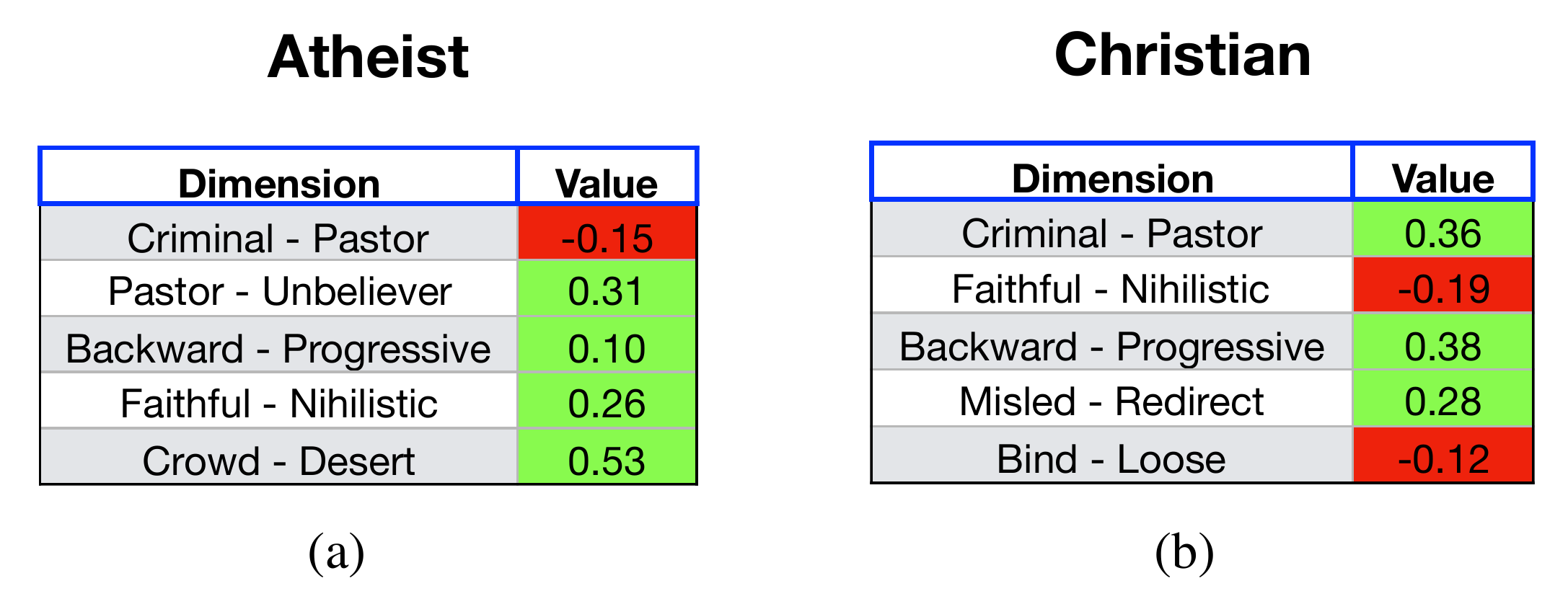}
	\caption{\emph{Explaining classification results.} We present the POLAR dimension as well as their corresponding value for the features that were assigned more weights by the classifier when classifying the articles as  (a) ``Atheist'' or (b) ``Christian''. Dimensions like ``criminal - pastor'', ``faithful - nihilist'' are assigned more weights. The selected dimensions also align well with human understanding.}
	\label{fig:expl_class}
\end{figure}

Apart from providing comparable performance across different downstream tasks, the inherent interpretability of POLAR dimensions also allows for explaining results of black box models.  
To illustrate, we consider the Religious news classification task and build a Random Forest model using word averaging as the feature. Utilizing the LIME framework~\cite{lime_2016}, we compare the sentences that were inferred as ``Christian'' by the classifier to those which were classified as ``Atheist''. In figure~\ref{fig:expl_class} we consider two such examples and report the dimensions/features (as well as their corresponding values) that were given higher weights in each case. Notably, the dimensions like `criminal - pastor', 'backward - progressive', `faithful - nihilistic' are given more weights for classification which also corroborates well with the human understanding. Note that the feature values across the POLAR dimensions which essentially represent projection to the polar opposite space, are relevant as well. For example, the article classified as ``Christian'' in figure~\ref{fig:expl_class} has a value $-0.15$ for the dimension `criminal - pastor' which means that it is more aligned to `pastor' while it is the opposite in case of `Atheist'. 
This demonstrates that POLAR dimensions could be used to help explain results in black box classification models. 
\begin{table}
	\centering
	\caption{\emph{Evaluation of the interpretability.} We report the conditional probability of the top $k$ dimensions as selected by the annotators to be among the ones selected by POLAR. We also report random chance probabilities for the selected dimension to be among the POLAR dimensions for different values of $k$. The probabilities for POLAR are significantly higher than the random chance probabilities indicating alignment with human judgement.}
	\begin{tabular}{llllll}
		\hline
		Top k            & 1 & 2 & 3 & 4 & 5  \\ \hline
		GloVe w\textbackslash{} POLAR  & 0.876 & 0.667 & 0.420 & 0.222 & 0.086  \\
		Random chance          & 0.5 & 0.22 & 0.083 & 0.023 & 0.005 \\
		\hline
	\end{tabular}
	\label{tab:human_judgement_topN}
\end{table}

\section{Discussion}

Finally, we discuss potential application domains for the presented POLAR framework, as well as limitations and further challenges.

\subsection{Applications}
In this paper we introduced POLAR, a framework for adding interpretability to pre-trained word embeddings. Through a set of experiments on different downstream tasks we demonstrated that one can add interpretability to existing word embeddings while mostly maintaining performance in downstream tasks. This should encourage further research in this direction. We also believe that our proposed framework could prove useful in a number of further areas of research. We discuss some of them below.

\noindent{\bf Explaining results of black box models.}
In this paper, we have demonstrated how the interpretable dimensions of POLAR are useful in explaining results of a random forest classifier. We believe POLAR could also be used for generating counterfactual explanations (~\citet{wachter2017counterfactual}) as well. Depending on the values of the POLAR dimensions one might be able to explain why a particular data point was assigned a particular label. 

\noindent{\bf Identifying bias.}
\citet{bolukbasi2016man} noted the presence of gender bias in word embeddings, giving the example of `Computer programmer' being more aligned to `man' and `Homemaker' being more aligned to `woman'. Preliminary results indicate that POLAR dimensions can assist in measuring such biases in embeddings. 
For example, the word `Nurse'  has a value of $-3.834$ in the dimension `man--women' which indicates that it is more strongly aligned with woman. We believe that POLAR dimensions might help identifying bias across multiple classes.

\noindent{\bf `Tunable' Recommendation.}
\citet{vig2011navigating} introduced a conversational recommender system which allows user to navigate from one item to other along dimensions represented by tags. Typically, given a movie, a user can find similar movies by tuning one or more tags (e.g., a movie like `Pulp fiction' but less dark). POLAR should allow for designing similar recommendation systems based on word embeddings in a more general way.

\subsection{Limitations}

\noindent{\bf Dependence of Interpretability on underlying corpora.} 
Although we demonstrated that the performance of POLAR on downstream tasks is similar to the original embeddings, its interpretability is highly dependent on the underlying corpora.
For example consider the word `Unlock'. The dimensions `Fee--Freebie' and `Power--Weakness' were selected by the human judges to be the most interpretable, while these two dimension were not present in the top 5 dimension of the POLAR framework. On closer examination, we observe that the top  dimensions of POLAR were not directly related to the word Unlock (`Foolish--Intelligent', `Mobile--Stationary', `Curve--Square', `Fool--Smart', `Innocent--Trouble'). We believe this was primarily due to the underlying corpora used to generate the baseline embeddings.

\noindent{\bf Identifying relevant polar opposites.}
Although we assume that the polar opposites are provided to POLAR by an oracle, selecting relevant polar opposites is critical to the performance of POLAR which can be challenging for smaller corpora. If antonym pairs are used as polar opposites, methods such as the ones introduced by \citet{an2018semaxis} could be used to find polar words as well as handling smaller corpora. 

\noindent{\bf Bias in underlying embeddings.}
Since, we use pre-trained embeddings, the biases present in them are also manifested in the POLAR embeddings as well. However, we believe the methods developed for removing bias, with minor modifications could be extended to POLAR as well.

\section{Conclusion and future direction}

We have presented a novel framework (POLAR) that adds interpretability to pre-trained embeddings without much loss of performance in downstream tasks. We utilized the concept of Semantic Differential from psychometrics to transform pre-trained word embeddings into interpretable word embeddings. The POLAR framework requires a set of polar opposites (e.g. antonym pairs) to be obtained from an oracle, and then identifies a corresponding subspace (the polar subspace) from the original embedding space. The original word vectors are then projected to this polar subspace to obtain new embeddings for which the dimensions are interpretable. 

To determine the effectiveness of our framework we considered several downstream tasks that utilize word embeddings, for which we systematically compared the performance of the original embeddings vs. POLAR embeddings. Across all tasks, we obtained competitive results. In some cases, POLAR embeddings even outperformed the original ones. 
We further performed human judgement experiments to determine the degree of interpretability of these embeddings. We observed that in most cases the dimensions deemed as most discriminative by POLAR aligned well with human understanding. 

\noindent{\bf Future directions.} An obvious next step would be to extend our framework to other languages as well as corpora. This would allow us to understand word contexts and biases across different cultures. We could also include other sets of polar opposites as well. Another interesting direction would be to investigate whether the POLAR framework could be applied to add interpretability to sentence and document embeddings which then might be utilized for explaining -- for example -- search results. 


\clearpage
\balance


\end{document}